\def\year{2020}\relax
\documentclass[letterpaper]{article} 
\usepackage{aaai20}  
\usepackage{times}  
\usepackage{booktabs}
\usepackage{helvet} 
\usepackage{courier}  
\usepackage{amsmath}
\usepackage{xcolor}
\usepackage{caption}
\usepackage{subcaption}
\usepackage{amssymb}
\usepackage[multiple]{footmisc}
\usepackage[hyphens]{url}  
\usepackage{graphicx}  
\title{Re-balancing Variational Autoencoder Loss for Molecule Sequence Generation}
\author{Chaochao Yan, Sheng Wang, Jinyu Yang, Tingyang Xu, Junzhou Huang \thanks{Corresponding author: J. Huang, jzhuang@uta.edu} \\
University of Texas at Arlington \\ 
Tencent AI Lab \\
}

\begin{document}

\maketitle

\begin{abstract}
Molecule generation is to design new molecules with specific chemical properties and further to optimize the desired chemical properties.
Following previous work, we encode molecules into continuous vectors in the latent space and then decode the vectors into molecules under the variational autoencoder (VAE) framework. 
We investigate the posterior collapse problem of current RNN-based VAEs for molecule sequence generation.
For the first time, we find that underestimated reconstruction loss leads to posterior collapse, and provide both theoretical and experimental evidence.
We propose an effective and efficient solution to fix the problem and avoid posterior collapse.
Without bells and whistles, our method achieves SOTA reconstruction accuracy and competitive validity on ZINC 250K dataset.
When generating 10,000 unique valid SMILES from random prior sampling, it costs JT-VAE 1450s  while our method only needs 9s. Our implementation will be made public. 
Our implementation is at \textcolor{blue}{\url{https://github.com/chaoyan1037/Re-balanced-VAE}}.

\end{abstract}

\noindent Discovering new molecules that have desired target properties is the key challenge of drug and material design. This can be considered as an optimization problem, and the goal is to search for molecules with the best desired property score \cite{gomez2018automatic}. However, exhaustive exploration in the molecule space is infeasible, as the number of estimated drug-like molecules is in the order of $ 10^{60} $ \cite{polishchuk2013estimation}. Additionally, molecule synthesis and validation are time-consuming and expensive in practice.

\begin{figure}[t]
\centering
  \includegraphics[width=0.48\textwidth]{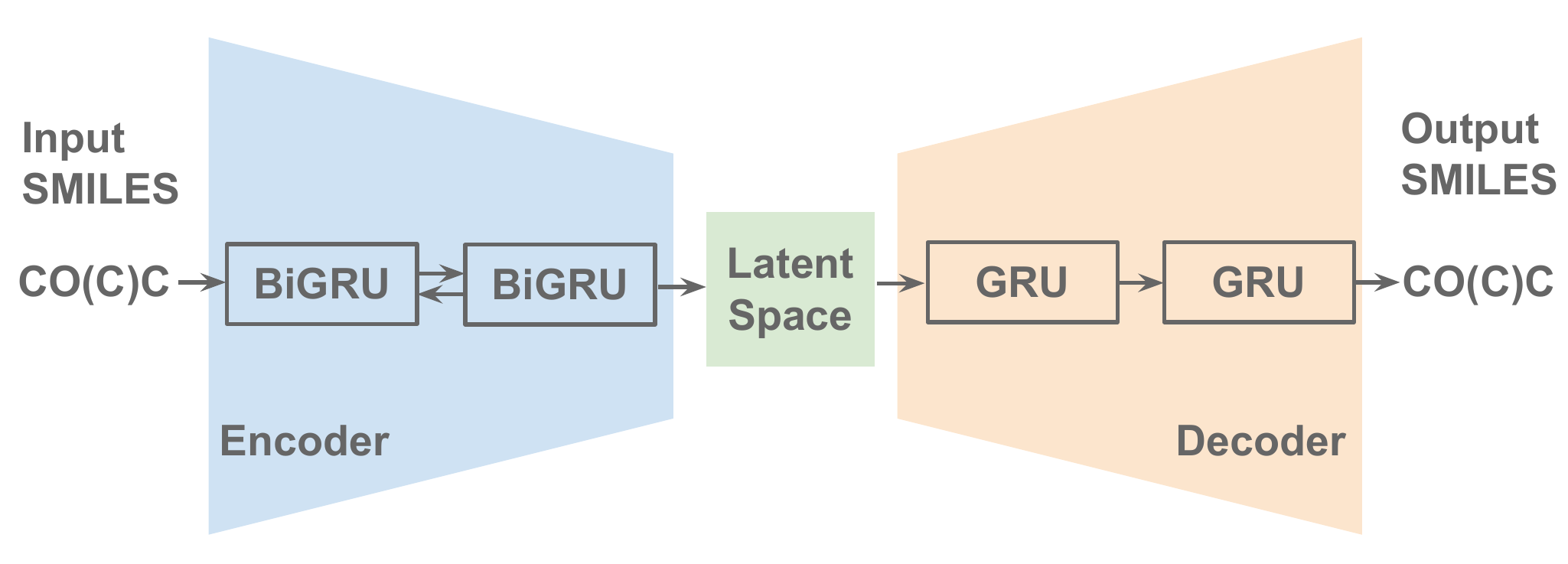}
  \caption{Overview of our VAE model. The encoder and docoder are built upon the bi-directional GRU and uni-directional GRU, respectively. Both input and output are SMILES sequences.}
  \label{fig:VAE}
\end{figure}

The majority of molecule generation methods heavily rely on the variational autoencoder (VAE) which is a combination of a deep latent variable model and an accompanying variational learning technique \cite{kingma2013auto} \cite{rezende2014stochastic}. As shown in Figure \ref{fig:VAE}, drug molecules can be first embedded by the encoder into the continuous latent space which can be further utilized for property prediction and optimization. After that, the decoder maps a continuous latent vector to reconstruct the input molecule. Thanks to the clustering ability of VAE, the latent representations of semantically similar molecules (with similar chemical structures and properties) are grouped together in latent space. In consequence, it allows semantically meaningful sampling and smooth interpolation in the latent space. Therefore, new molecules can be generated by randomly sampling from the prior and can be further optimized by exploring the latent space. The key idea behind the optimization above is to search for molecules that maximize an property score objective, given molecules' latent representation as input\cite{gomez2018automatic}. 

However, existing VAE models mainly suffer from the posterior collapse issue, where the decoder tends to ignore the latent vectors \cite{bowman2016generating} \cite{gomez2018automatic}. This problem is more frequently observed in those models with RNN-based backbone \cite{he2019lagging}. As a consequence, the generated molecules tend to be in low diversity and are weakly relevant to the latent vectors \cite{gomez2018automatic} \cite{kusner2017grammar}. This phenomenon has also been observed in Natural Language Processing (NLP) tasks, such as text generation \cite{bowman2016generating}. To alleviate this problem, the major focus of the previous studies is to adopt various training strategies, such as KL cost annealing \cite{bowman2016generating} or aggressively optimizing the decoder before each encoder update \cite{he2019lagging}. However, such methods can not be simply extended to molecule generation, mainly stemming from the fact that the molecule sequences are strictly structured and any mutations can result in invalid sequences. Motivated by the success of parse trees in the NLP field and attribute grammars in compiler design, recent work \cite{kusner2017grammar} \cite{dai2018syntax} incorporate grammar or syntax rules to guarantee that syntactically valid SMILES sequences can be generated. As an alternative, a molecule can also be represented by a graph in order to mitigate the posterior collapse \cite{li2018learning} \cite{jin2018junction}.

Inspired by the essential pitfalls of the contemporary RNN-based VAE models in the molecule generation, here, we propose a new method to alleviate the posterior collapse issue. To achieve this goal, we first analyze the posterior collapse of vanilla VAE model for SMILES sequence generation. For the first time, we find that the posterior collapse is largely triggered by the underestimated reconstruction loss. We, therefore, propose to use a novel loss function to leverage the trade-off between the reconstruction loss and the KL loss in the VAE training. Without making any changes on the VAE network structures or introducing additional computational complexity, our method is extremely simple yet effective in preventing posterior collapse. We also provide the theoretical analyse of our method, and empirically demonstrate its state-of-the-art (SOTA) reconstruction accuracy and competitive validity score on the ZINC 250K dataset. Our primary contributions can be summarized as:

\begin{itemize}
\item  We diagnose the main reason causing the posterior collapse within the RNN-based VAE model for molecule generation, with both theoretical and intuitive analyses been provided.  
\item We propose an effective and efficient method to eliminate the posterior collapse in VAE by leveraging the associations between the reconstruction loss and the KL loss. 
\item Extensive empirical studies demonstrate our method's superiority over SOTA molecule generation approaches on the ZINC 250K dataset.
\end{itemize}

\section{Background Information}
\subsection{The Variational Autoencoder}
The VAE \cite{kingma2013auto} \cite{rezende2014stochastic} is a specially regularized version of the standard autoencoder (AE). It is appealing because it can learn complex distribution in an unsupervised manner and later act as a generative model defined by a prior $p(z)$ and a conditional distribution $p_{\theta}(x|z)$. Since the true data likelihood is usually intractable, so the VAE instead optimizes an evidence lower bound (ELBO) which is a valid lower bound of the true data log likelihood:

\begin{equation} \label{ELBO}
\begin{split}
\mathcal{L}(x; \theta, \phi) & =  \mathbb{E}_{q_{\phi}(z|x)}[\log p_{\theta}(x|z)] - D_{\text{KL}}(q_{\phi}(z|x) || p(z)) \\
& \leq \log p(x).
\end{split}
\end{equation}
where the encoder $q_{\phi}(z|x)$ is parameterized with $\phi$ and learns to map the input $x$ to a variational distribution, and the decoder $p_{\theta}(x|z)$ parameterized with $\theta$ tries to reconstruct the input $x$ given the latent vector $z$ from the learned distribution. Usually, $q_{\phi}(z|x)$ is modeled as a Gaussian distribution and  optimized to approximate the true posterior $p_{\theta}(z|x)$. 

The VAE is optimized to maximize ELBO (\ref{ELBO}), where (i) negative reconstruction loss $ \mathbb{E}_{q_{\phi}(z|x)}[\log p_{\theta}(x|z)]$ enforces the encoder to generate meaningful latent vector $z$, so that the decoder can reconstruct the input $x$ from the $z$, and (ii) the KL regularization loss $D_{\text{KL}}(q_{\phi}(z|x) || p(z))$ minimizes the KL divergence between the approximate
posterior $ q_{\phi}(z|x) $ and the prior $p(z) \sim \mathcal{N}(0, \mathbf{I})$.

\section{Related Work}
\subsection{Text Generation with VAE}
Motivated by the ubiquitous posterior collapse problems observed in VAE-based models for text generation, various methods have been proposed and investigated recently \cite{bowman2016generating} \cite{yang2017improved} \cite{higgins2017beta} \cite{kim2018semi}. Bowman et al. \cite{bowman2016generating} propose to anneal the KL loss weight to enable the model to learn meaningful encoding before applying KL loss to cluster the encoding. They also weaken the decoder with word dropout and historyless decoding to force the decoder rely on the latent vectors. 
\cite{he2019lagging} concludes that the posterior collapse mainly attributes to the lagging encoder network's inability in approximating the true posterior. To overcome this limitation, they propose a novel training strategy which aggressively optimize the encoder network. Inspired by \cite{bowman2016generating}, \cite{fu2019cyclical} proposes a cyclical annealing strategy which repetitively starts training from a pretrained model resulted in the previous cycle. They claim this procedure can make the model learn more meaningful latent representations progressively.

\subsection{Molecule Generation}
Thanks to the development of NLP text generation, the VAE model is applied for molecule generation for the first time in CVAE \cite{gomez2018automatic}. They build a VAE encoder and decoder with GRU layers, representing molecules in the SMILES sequences. However, their model suffers from generating invalid SMILES sequences which makes their model impracticable. To improve the prior validity, context-free grammars for SMILES are introduced in GVAE \cite{kusner2017grammar} to represent a molecule in the sparse tree. However, the validity score is still unsatisfactory. Inspired by this method, Syntax-directed VAE (SD-VAE) \cite{dai2018syntax} incorporates extra semantic rules to ensure generated SMILES valid, and it achieves the best performance among all SMILES-based methods. However, these models did not solve model posterior collapse problem and there is a large space to improve.

Except for SMILES representations, molecules can also be represented in graph. \cite{li2018learning} employs graph-structured representations for molecules and models the probabilistic dependencies among a graph’s nodes and edge with graph neural networks. Molecules are generated node by node by their model. Chemical sub-graphs instead of atoms are used as the basic building blocks in JT-VAE \cite{jin2018junction}, their methods can incrementally generate molecules to ensure chemical validity at each step. JT-VAE is the SOTA model for molecule generation.

\section{Problem and Solution}

\subsection{Posterior Collapse}
Prior work \cite{bowman2016generating} \cite{yang2017improved} \cite{higgins2017beta} \cite{kim2018semi} on NLP text generation has observed the posterior collapse phenomenon, in which the decoder tends to ignore $z$ when training the VAE model. When posterior collapse happens, the model training falls into the the local optimum of the ELBO objective (\ref{ELBO}), in which the variational posterior $q_{\phi}(z|x)$ naively mimics the model prior $p(z)$. Note that the KL loss in ELBO can be further decomposed \cite{hoffman2016elbo} as:

\begin{equation} \label{KL_mutual_info}
\begin{split}
 \mathbb{E}_{p_d{(x)}}[D_{\text{KL}}(q_{\phi}(z|x) || p_(z))] =  &I_q + \\
& D_{\text{KL}}(q_{\phi}(z) || p(z)),
\end{split}
\end{equation}
where $I_q$ is the mutual information between $x$ and $z$ given $q_{\phi}(z|x)$, and $p_d(x)$ is empirical data distribution. 
When posterior collapse occurs, the KL loss decreases nearly to zero so that $I_q$ is also close to zero (both items on the right-hand side in (\ref{KL_mutual_info}) are non-negative) during the VAE model training process. It is especially evident when modelling discrete data with a strong auto-regressive network such as LSTM \cite{hochreiter1997long} and GRU \cite{chung2014empirical}, which is exactly our case. This is undesirable since the VAE model fails to learn meaningful latent representations for input sequences.

For NLP text generation task, the posterior collapse problem has been mainly attributed to the low quality of latent representations $z$ at the early stage of model training \cite{bowman2016generating} \cite{he2019lagging} \cite{fu2019cyclical}. To be more specific, the decoder $p_{\theta}(x|z)$ falls behind the encoder $q_{\phi}(z|x)$ at the initial training procedure, and $q_{\phi}(z|x)$ generates low-quality latent representations so that it is very hard for $p_{\theta}(x|z)$ to recover the input sequences. In consequence, the model is forced to ignore $z$. Many solutions have been proposed to solve the problem and they have demonstrated satisfactory improvement on NLP datasets. 

However, the molecule generation is a quite different scenario though it appears to be same as the NLP text generation. First of all, its token size is far more less than the NLP text generation. The token size for NLP text is usually tens of thousands or even more, while it is less than 100 for chemical molecule data. The smaller token size makes the molecule reconstruction task much easier. Second, the molecule sequence is composed strictly following the SMILES grammar or syntax rules, and the reconstructed sequence must be exactly the same as the input to be matched. Any token mutations can result in a completely different sequence. However, there are no rigid grammar rules applied to the NLP text and exact match is not required. 

We have found existing solutions \cite{he2019lagging} \cite{fu2019cyclical} to posterior collapse in NLP text generation does not work well for chemical molecule generation. This motivates us to propose such a solution for molecule generation.

\subsection{The Problem in Previous Solutions}
To avoid posterior collapse, which will cause a VAE losing reconstruction ability, previous SMILES-based methods CVAE \cite{gomez2018automatic}, GVAE \cite{kusner2017grammar}, and SD-VAE \cite{dai2018syntax} reduce  the standard  deviation $\sigma$ of prior Gaussian distribution to a small value 0.01 (can be found in their public implementation CVAE\footnote{https://github.com/aspuru-guzik-group/chemical$\_$vae}\footnote{https://github.com/HIPS/molecule-autoencoder}, GVAE\footnote{https://github.com/mkusner/grammarVAE}, SD-VAE\footnote{https://github.com/Hanjun-Dai/sdvae}),  which makes their models more like AEs instead of VAEs. That is why CVAE and GVAE have a decent reconstruction accuracy but extremely low validity scores as shown in Table \ref{tab:reconstruction-validity}. If we set the $\sigma$=1, all these three models will suffer from model posterior collapse and lose the reconstruct ability (similar to the vanilla VAE in Figure \ref{fig:vae_traing_dynamic}(e)). In following our analysis and experiments, we strictly keep the $\sigma$=1.

\subsection{Underestimated Reconstruction Loss}
To investigate the cause of posterior collapse within the VAE for molecule generation, we conduct convincing analysis and investigation into posterior collapse. We hypothesize it is the underestimated reconstruction loss that causes posterior collapse of a VAE during training process. Both theoretical analysis and experimental support are provided.

From the perspective of theory, reconstruction loss term $  \mathbb{E}_{q_{\phi}(z|x)}[\log p_{\theta}(x|z)]$ measures the reconstruction ability of the decoder given latent vector $z$. The decoder should only receive information from $z$ and tries to reconstruct the full sequence accurately from the given starting $z$.
However, in practice RNN model is usually optimized with teacher forcing \cite{williams1989learning}, in which the current input is the ground truth instead of prediction from a prior time step. 

We can rewrite the reconstruction loss term  in (\ref{ELBO}) as:
\begin{equation} \label{reconstruction_loss_old}
 \mathbb{E}_{q_{\phi}(z|x)}[ \sum_{t=1}^{T}{  \log p_{\theta}(x_t|z, \tilde{x}_{<t} ) }] ,
\end{equation}
where the $T$ is the maximum time step,
$\tilde{x}_{<t}$ is the prediction prefix before time $t$ and the current input is the previous time step output $\tilde{x}_{t-1}$, and $\tilde{x}_{0}$ is the start symbol.

With teacher forcing, the actual reconstruction loss is:
\begin{equation} \label{reconstruction_loss_actual}
 \mathbb{E}_{q_{\phi}(z|x)}[ \sum_{t=1}^{T}{  \log p_{\theta}(x_t|z, \tilde{x}_{<t},  x_{<t}) }] ,
\end{equation}
where $ x_{<t}$ is the ground-truth prefix before time $t$ and the ground-truth token of previous time step $ x_{t-1}$ is the RNN input, and ${x}_{0}$ is also the start symbol.

Since the ground-truth information is incorporated additionally in (\ref{reconstruction_loss_actual}) when training the VAE, which can make the prediction easier since the ground-truth prefix is given, we can expect that the reconstruction ability of decoder is largely overestimated compared with (\ref{reconstruction_loss_old}).  Therefore, we can assume the reconstruction loss term is underestimated, which will potentially breaks the balance between reconstruction loss and KL loss in (\ref{ELBO}). We will verify the assumption and also demonstrate quantitatively how much the reconstruction loss is underestimated in the experiment section. Let us agree on the claim for now.

\subsection{Re-balanced VAE Loss}
Since reconstruction loss is underestimated, and it breaks the balance with KL loss, which leads to the posterior collapse finally. We can recover the balance by applying a reconstruction loss weight $\alpha$ to the ELBO (\ref{ELBO}):
\begin{equation} \label{alpha}
\begin{split}
\mathcal{L}(x; \theta, \phi) = & \alpha \mathbb{E}_{q_{\phi}(z|x)}[\log p_{\theta}(x|z)] \\
& - D_{\text{KL}}(q_{\phi}(z|x) || p(z)), \alpha > 1,
\end{split}
\end{equation}
where $\alpha$ can be estimated using Monte Carlo in every training iteration. Specifically, we can sample a batch of data as input and run a VAE with/without teacher forcing, respectively. Since the reconstruction loss without teacher forcing can be regarded as the ``true" reconstruction loss, we approximate $\alpha$ as the ratio of reconstruction loss without teacher forcing to that with teacher forcing. However, estimating $\alpha$ in every training iteration is too expensive. We can set $\alpha$ as a hype-parameter for simplicity. 

Inspired by the $\beta$-VAE \cite{higgins2017beta} formulation, we can instead reduce KL loss weight $\beta$, which is equivalent to increasing reconstruction loss weight $\alpha$. It is more natural and convenient to search for the optimal value of hype-parameter $\beta$ since increasing $\beta$ from 0 is a gradual transition from AE to VAE. So we can have a similarly modified VAE loss formulation:
\begin{equation} \label{beta-vae}
\begin{split}
\mathcal{L}(x; \theta, \phi) =& \mathbb{E}_{q_{\phi}(z|x)}[\log p_{\theta}(x|z)] \\
& - \beta  D_{\text{KL}}(q_{\phi}(z|x) || p(z)), 0 \leq \beta < 1.
\end{split}
\end{equation}

Note that in our case $\beta < 1$, while $\beta$-VAE requires the KL weight $\beta > 1$. $\beta$-VAE is proposed in \cite{higgins2017beta} to learn disentangled representation of generative factors by enforcing a larger penalty on KL loss, since they postulate that $\beta > 1$ could place a stronger constraint on the latent representation to drive the VAE to learn more efficient latent representation of input $x$. While we have a completely different motivation and goal of fixing imbalanced VAE loss by reducing KL weight since we find reconstruction loss is underestimated in ELBO (\ref{ELBO}).

Except for theoretical analysis, our method can also be explained from an intuitive perspective. In previous methods CVAE, GVAE, and SD-VAE, when sampling latent vectors $z$ they have to reduce the standard deviation $\sigma$ to a small value 0.01 otherwise the model will collapse and lose the reconstruct ability.  Instead of reducing sampling $\sigma$, we can anneal the KL loss weight $\beta$ to make the model transform from AE to VAE gradually \cite{bowman2016generating}. Different from \cite{bowman2016generating}, we restrict $\beta$ to be smaller than 1. By searching for the optimal $\beta$, we can arrive a trade-off between the reconstruction accuracy and validity score.

We acknowledge that previous methods have empirically tried to reduce the KL loss weight to avoid the posterior collapse \cite{dai2018syntax} \cite{he2019lagging}  \cite{fu2019cyclical}. $\beta$-VAE ($\beta$ = 0.4) alleviates the problem and achieves competitive performance on density estimation for NLP text datasets \cite{he2019lagging}, which proves that reducing $\beta$ is viable for NLP text task. It is also indicated that setting $\beta = 1 / \mathbf{LatentDimension} $ could lead to better results \cite{kusner2017grammar} \cite{dai2018syntax}. But none of these methods provided any analysis or explanation, they are completely empirical. We are the first to recognize the underestimated reconstruction loss leads to posterior collapse problem, and further we officially propose to reduce KL loss weight to overcome the posterior collapse with solid support, both theoretically and intuitively.

\begin{figure*}[t]
\centering
\includegraphics[width=0.84\textwidth]{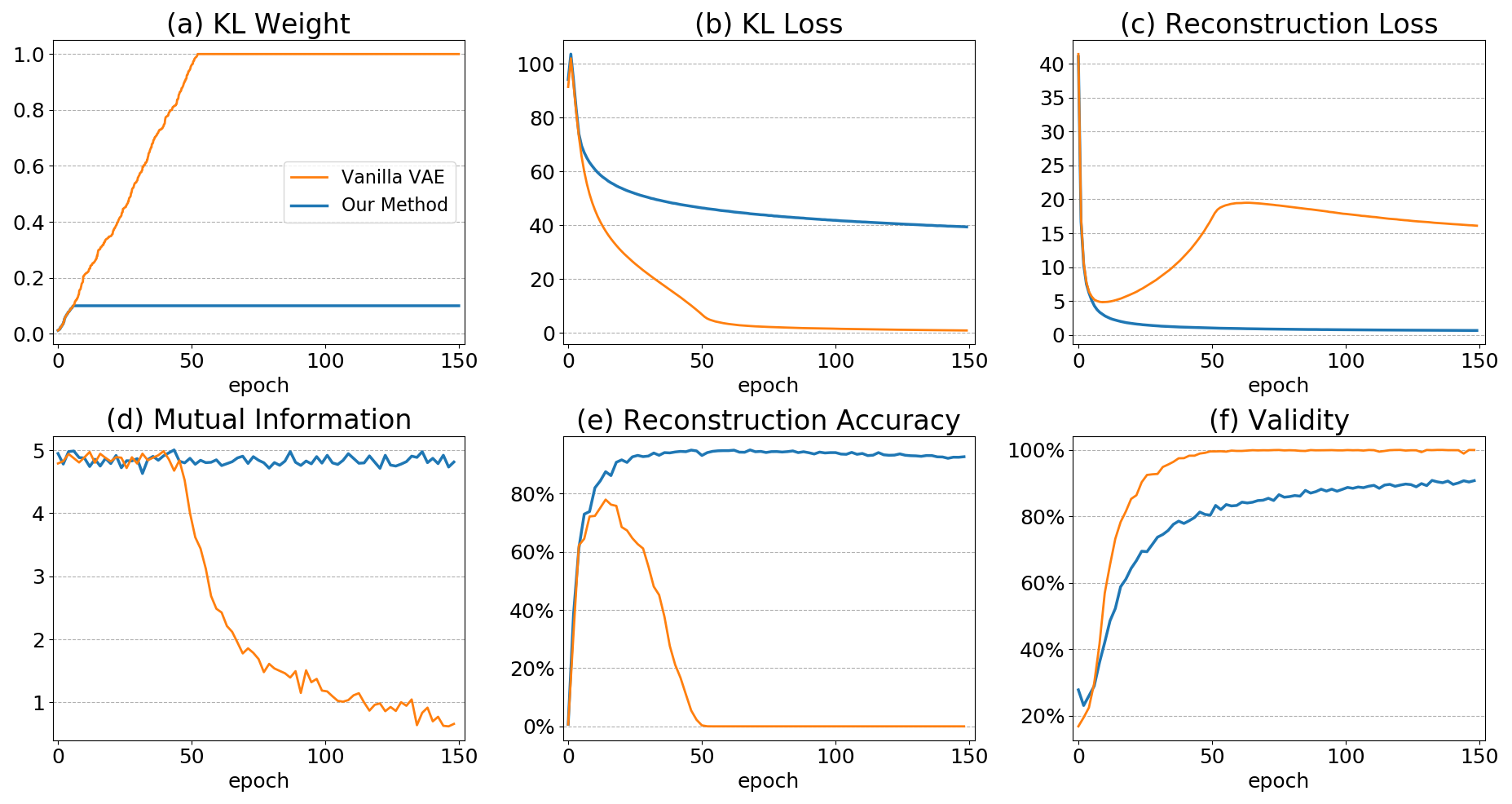} 
\caption{Training dynamic of vanilla VAE model on validation data. We track 
(a) KL weight $\beta$,
(b) KL loss $D_{\text{KL}}(q_{\phi}(z|x) || p(z))$, 
(c) reconstruction loss  $-\mathbb{E}_{q_{\phi}(z|x)}[\log p_{\theta}(x|z)]$,
(d) mutual information $I_q$,
(e) reconstruction accuracy,
and (f) validity during the training.
The orange line is the vanilla VAE with training KL loss annealing, and the maximum KL weight $\beta$ is 1. Our method (Blue) reduces the maximum value of $\beta$ to 0.1. Both models are trained with KL weight annealing and teacher forcing.}
\label{fig:vae_traing_dynamic}
\end{figure*}

\section{Experiments}
Our proposed solution to the VAE model posterior collapse is simple but extremely effective and efficient. We do not need modify the network architecture and only adjust the training loss slightly, without introducing much extra computation. In this section, we will first train a vanilla VAE model and track the occur of model collapse, as well as experimentally verify that the reconstruction loss is underestimated. Then we will conduct extensive experiments to demonstrate the effectiveness of our proposed method.

\subsection{VAE Architecture, Dataset, and Evaluation Metrics}
We build our VAE model based on GRU. The VAE encoder is composed of two layers of bi-directional GRU which is better at capturing the sequence representation \cite{schuster1997bidirectional}, and the hidden size of each layer is 512. The decoder is made up of four layers of uni-directional GRU with the same hidden size 512. Following previous work \cite{gomez2018automatic} \cite{jin2018junction}, we use unit Gaussian prior and set the latent vector dimension to be 56. The ELBO objective is optimized with Adam \cite{kingma2014adam} and learning rate is 0.0001.The model is trained with teacher forcing and KL loss annealing following previous work. Since the model has a really good convergence, we train the model for 150 epochs and report the performance of the final model. We implement our model using PyTorch \cite{paszke2017automatic}. Experiments are conducted on a machine with a  Intel Core i7-5930K@3.50GHz CPU and a GTX 1080 Ti GPU.

We conduct all our experiments on ZINC 250K dataset \cite{kusner2017grammar} which is a subset of the ZINC \cite{sterling2015zinc}. Molecule sequences are tokenized with the regular expression from \cite{schwaller2018found}. We use the same training and testing split as previous work \cite{kusner2017grammar} \cite{jin2018junction}, and have 10K hold-out data out of the training as the validation data. From now on, we will use the same experimental setting in all our experiments unless explicitly stated. 

As for the model evaluation metrics, we report the reconstruction accuracy and validity score like previous work.  Following \cite{jin2018junction}, we encode each molecule from test dataset 10 times, and then decode 10 times for each latent vector.  The reconstruction accuracy is defined to be the ratio of successfully reconstructed molecule sequences to the total tried reconstruction. The  reconstructed SMILES must be exactly the same as the input to be counted as successful. To calculate validity, 1000 latent vectors are randomly sampled from the prior distribution, and each is decoded 100 times. 
The validity is the portion of chemically valid reconstruction SMILES to the total decoded sequences. We use RDkit \cite{landrum2006rdkit} to check if a SMILES is valid.

\subsection{VAE Training Dynamic}
We track the training process of a vanilla VAE model for SMILES sequences, as well as that of our proposed method. By investigating the training dynamic like KL weight, KL loss, reconstruction loss, mutual information, as well as the model performance (reconstruction accuracy and validity), we conclude that the underestimated reconstruction loss causes the posterior collapse of vanilla VAE model during the training process. Mutual information $I_q$ can be calculated using Monte Carlo sampling as proposed in \cite{hoffman2016elbo} \cite{dieng2019avoiding}:
\begin{equation} \label{KL_mutual_info_monte_carlo}
\begin{split}
 I_q = & \mathbb{E}_{p_d{(x)}}[D_{\text{KL}}(q_{\phi}(z|x) || p_(z))] - \\
& D_{\text{KL}}(q_{\phi}(z) || p(z)) ,
\end{split}
\end{equation}
which is actually the same as the (\ref{KL_mutual_info}). We approximate the aggregated posterior $ q_{\phi}(z) = \mathbb{E}_{p_d{(x)}}[q_{\phi}(z|x)] $ using Monte Carlo sampling. $ D_{\text{KL}}(q_{\phi}(z) || p_(z)) $ can also be estimated by the Monte Carlo, and we can obtain samples from $q_{\phi}(z)$ by ancestral sampling. More details about $I_q$ computation can be found in \cite{hoffman2016elbo}.

As a comparison, we also illustrate the training dynamic when our proposed method is applied. For our method, we set the KL weight $\beta$ = 0.1 which is the optimal parameter we found. We keep all the other experimental settings the same as the vanilla VAE to make a fair comparison.

Results of two models run are plot in the Figure \ref{fig:vae_traing_dynamic}. The vanilla VAE model performances well on the validation data at the early stage of the KL weight annealing. As the KL weight increases, KL loss drops quickly as expected since more penalty is added to the KL loss term, while the small reconstruction loss starts to rise at the same time. The mutual information $I_q$ decreases to 0.65 at the end, which means the decoder does not absorb much information from the latent vectors when generating the output. This evidence indicates the posterior collapse has happened. When looking at the model performance on validation data, we can notice that the reconstruction accuracy is close 0\% while the validity score is almost perfect. This indicates that too much pressure has been placed on the KL loss, which breaks the balance between the reconstruction loss and KL loss and results in the model posterior collapse. 

Our method achieves lower reconstruction loss early and can maintain it during model training. Although the KL loss of our method is larger than the vanilla VAE,  considering that we have a much smaller KL weight $\beta$ now, the equivalent KL loss added to the training objective should still be in the normal range. Especially, our method maintains the mutual information to be around 4.8, which means output sequences are strongly related to latent vectors. As for the model performance, our method achieves 92.7\% reconstruction accuracy and 90.7\% validity score, which proves the superiority of our method. 

\subsection{Proof of Underestimated Reconstruction Loss}

\begin{figure}[!htb]
\centering
  \includegraphics[width=0.45\textwidth]{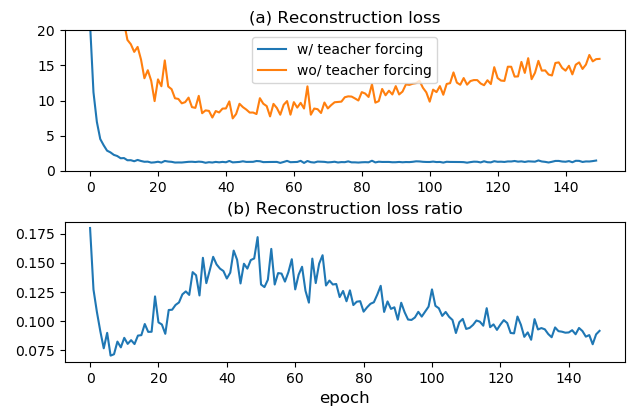}
  \caption{(a) Reconstruction loss on validation dataset. 
  At each time step, models parameters are the same when calculating training and testing loss. (b) Reconstruction loss underestimated ratio.}
  \label{fig:KL_loss}
\end{figure}

We hypothesize that introducing ground-truth information into the decoder will result in underestimated reconstruction loss, and have provided our detailed analysis previously. In this section, we will experimentally verify that the reconstruction loss is indeed underestimated during the training. 
We can estimate how much the reconstruction loss has been underestimated using Monte Carlo Sampling. Specifically, we can sample a batch of data, then run the model with and without the teacher forcing, respectively. The underestimated ratio can be approximated by the ratio of reconstruction loss with teacher forcing to that without teacher forcing.

We track the reconstruction loss on the validation dataset when the teacher forcing is applied and removed, respectively. Results are shown in the Figure \ref{fig:KL_loss}(a). When teacher forcing is applied, the reconstruction loss drops close to 1 quickly, while the loss is much larger (at least 7.5) without teacher forcing. This is expected since without teacher forcing, any wrong prediction token as input may result in following prediction totally different from ground-truth sequences.

To figure out how much the reconstruction loss has been underestimated, we can compute the ratio as reconstruction loss w/ teacher forcing to that wo/  teacher forcing  at each time step. Results are shown in Figure \ref{fig:KL_loss}(b). It confirms our claim that the reconstruction loss is underestimated. To recover a balanced VAE loss, we can set KL loss weight exactly as the underestimated ratio in each epoch. To be simplified, we set $\beta=0.1$ and we find it works well in practice.

\begin{table}[!htb]
\caption{Reconstruction accuracy and validity results. Baseline results are reported in \cite{kusner2017grammar} \cite{dai2018syntax} \cite{simonovsky2018graphvae} \cite{jin2018junction}.}
\centering
{
\begin{tabular}{ p{2cm} p{2cm}  p{2cm} } 
 \hline
 \textbf{Model} & \multicolumn{1}{c}{\textbf{Reconstruction}} & \multicolumn{1}{c}{\textbf{Validity}} \\ 
 \hline
 &\multicolumn{1}{c}{\textbf{SMILES-based}}& \\[0.4ex]
 CVAE & \multicolumn{1}{c}{44.6\%} & \multicolumn{1}{c}{0.7\%} \\
 GVAE & \multicolumn{1}{c}{53.7\%} & \multicolumn{1}{c}{7.2\%} \\ 
 SD-VAE & \multicolumn{1}{c}{76.2\%} & \multicolumn{1}{c}{43.5\%} \\
 Our Method & \multicolumn{1}{c}{\textbf{92.7}\%} & \multicolumn{1}{c}{90.7\%} \\
 \hline
 &\multicolumn{1}{c}{\textbf{Graph-based}} &\\[0.4ex]
 GraphVAE & \multicolumn{1}{c}{-} & \multicolumn{1}{c}{13.5\%} \\
 JT-VAE & \multicolumn{1}{c}{76.7\%} & \multicolumn{1}{c}{\textbf{100.0}\%} \\
 \hline
\end{tabular}
}
\label{tab:reconstruction-validity}
\end{table}

\subsection{Molecule Reconstruction Accuracy and Validity}

We summarize the molecule reconstruction accuracy and validity on test dataset in the Table \ref{tab:reconstruction-validity}. Our method outperforms all previous models in reconstruction accuracy by a large margin (16\% larger than the second best model). In the meanwhile, our method achieves 90.7 \% validity, which is the second best among all the models.

Compared with other SMILES-based methods, our model is much more superior in both the reconstruction accuracy and prior validity, even if complex grammar or syntax rules are incorporated \cite{kusner2017grammar} \cite{dai2018syntax}. Note that JT-VAE model assembles molecules by adding sub-graphs step-by-step to make sure the generated molecule graphs are always valid. However, the sub-graphs are extracted from the training dataset, which limits the JT-VAE can not generate molecules with unseen sub-graphs. Our method achieves competitive validity performance without any constraints, and is able to generate novel molecules that are not from the same distribution as the training data. That is one important reason why our method achieves the STOA reconstruction accuracy, while JT-VAE suffers from reconstructing testing molecules \cite{mohammadi2019penalized}. Besides, our method is much more efficient than JT-VAE. When generating 10,000 unique valid SMILES from prior random sampling, JT-VAE\footnote{https://github.com/wengong-jin/icml18-jtnn}(faster version) takes about 1450s while our method only needs 9s.

\begin{figure}[!htb]
\centering
  \includegraphics[width=0.4\textwidth]{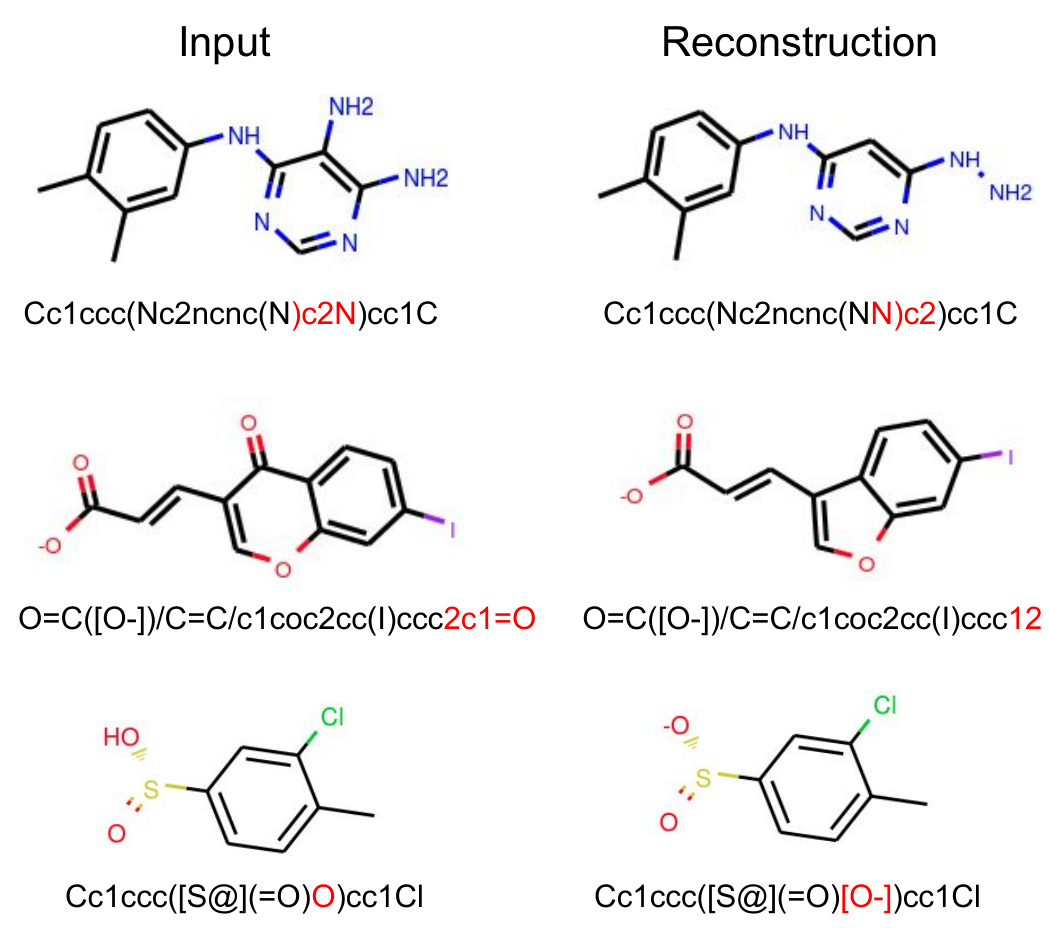}
  \caption{Reconstruction error examples. Unmatched tokens between the input and reconstruction SMILES are are shown in red (``[O-]" is one token).
  }
  \label{fig:miss_reconstruction}
\end{figure}
\begin{figure*}[t]
\centering
  \includegraphics[width=0.8\textwidth]{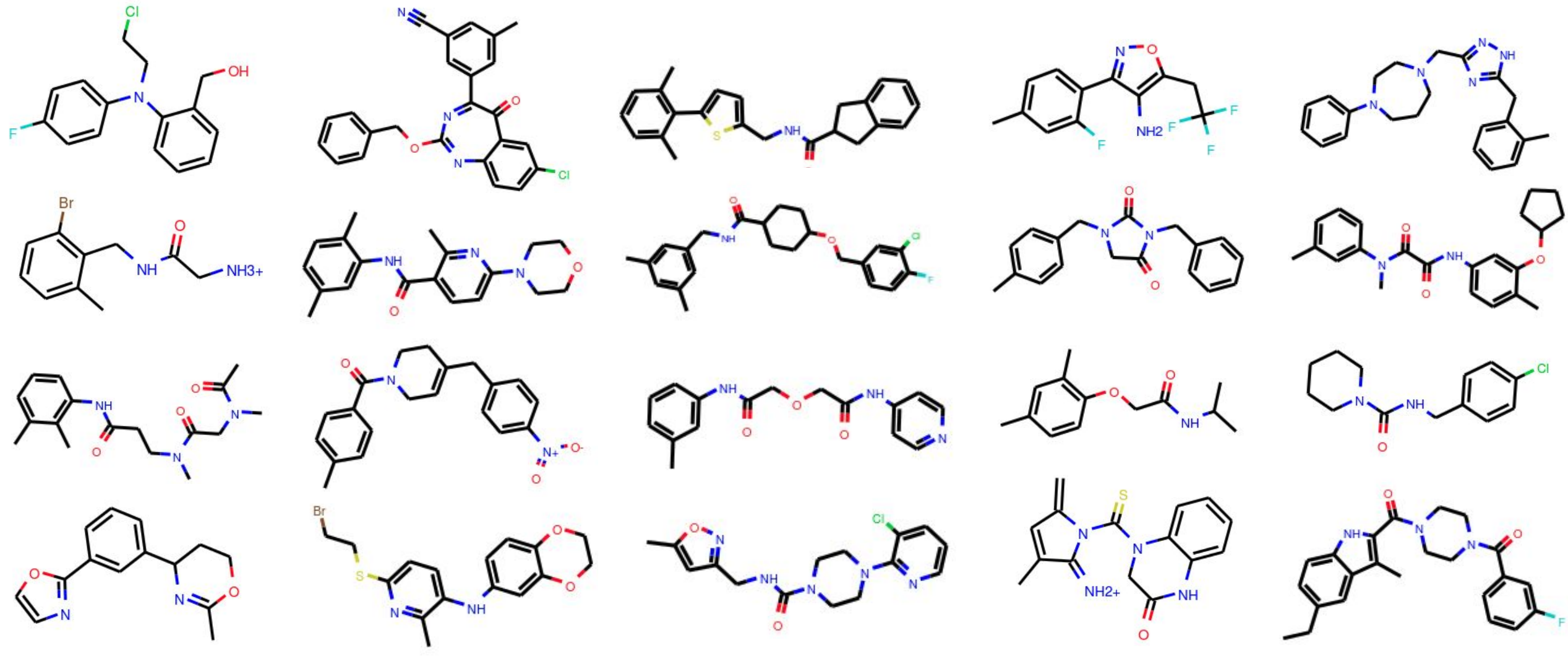}
  \caption{Generated molecules by random sampling from the prior.}
  \label{fig:prior_samples_max_01_latent_56}
\end{figure*}

\subsection{Error Analysis and Visualization}

Our model achieves 92.7\% reconstruction accuracy. We investigate the reconstruction results further and find that our model can predict 97.3\% of all tokens correctly, which is measured on the level of token instead of the sequence. Besides, most of unmatched sequences (62\%) are valid. These evidences indicate that our model is very well learned. We show some valid but unmatched examples in Figure \ref{fig:miss_reconstruction}.

As for the validity, we also investigate model outputs. We illustrate some generated molecules in the Figure \ref{fig:prior_samples_max_01_latent_56}, which demonstrates that our model can generate complicated and diverse molecules with multiple rings. 

For those invalid sequences, from both the reconstruction and prior sampling, there are several typical errors: (1) unkekulized atoms, (2) valence error, (3) unclosed ring, and (4) parentheses error. We believe that advanced techniques like grammar and syntax rules \cite{kusner2017grammar} \cite{dai2018syntax} are necessary and can help to reduce these kind of errors, and our method is essential and complementary to these methods.

\subsection{Bayesian Optimization}
One of the important tasks in the drug molecule generation is to make molecules with desired chemical properties. We follow \cite{kusner2017grammar} \cite{jin2018junction} for all the experimental setting, and the optimization target score is:
\begin{equation} \label{bayesian_optimization_target}
 y(m) = logP(m) - SA(m) - cycle(m),
\end{equation}
where $logP(m)$ is the octanol-water partition coefficients of meolecule $m$, $SA(m)$ is synthetic accessibility score, and $cycle(m)$ is number of large rings with more than six atoms.

\begin{table}[!htb]
\caption{Top-3 molecule property scores found by the BO. Baseline results are copied from  \cite{kusner2017grammar} \cite{dai2018syntax} \cite{jin2018junction} \cite{jin2018junction}.}
\centering
\begin{tabular}{p{2cm} p{1.2cm} p{1.2cm} p{1.2cm} } 
\hline
 \textbf{Model} & \textbf{1st} & \textbf{2nd} & \textbf{3rd}\\ 
 \hline
 \multicolumn{4}{c}{\textbf{SMILES-based}} \\[0.4ex]
 CVAE & 1.98 & 1.42 & 1.19 \\
 GVAE & 2.94 & 2.89 & 2.80 \\ 
 SD-VAE & 4.04 & 3.50 & 2.96 \\
  Our Method & \textbf{5.32} & \textbf{5.28} & \textbf{5.23} \\
 \hline
 \multicolumn{4}{c}{\textbf{Graph-based}} \\[0.4ex]
 JT-VAE & 5.30 & 4.93 & 4.49  \\
 \hline
\end{tabular}
\label{tab:BO}
\end{table}

\begin{figure}[!htb]
\centering
  \includegraphics[width=0.4\textwidth]{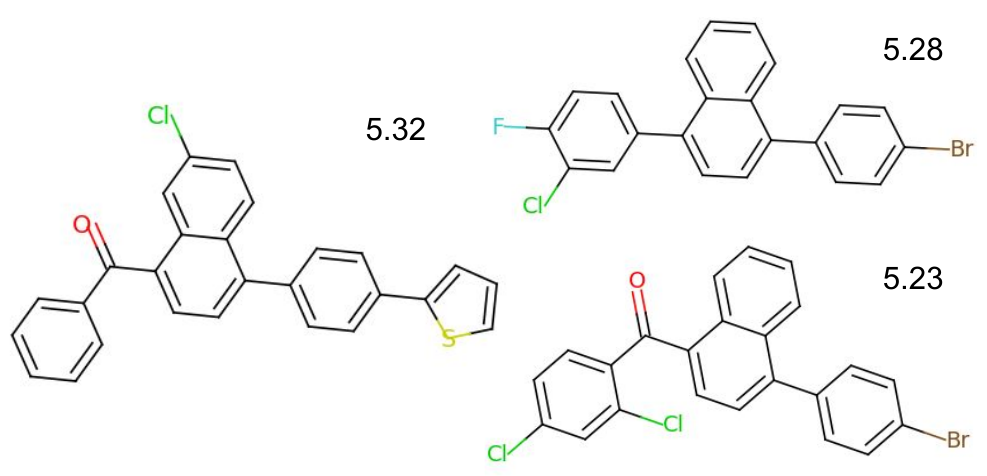}
  \caption{Top-3 molecules and associated scores found by our model with Bayesian optimization.}
  \label{fig:top3bo}
\end{figure}

We first associate each molecule with a latent vector which is the mean of the learned variational encoding distribution. The latent vector for each molecule will be treated as its feature and we train a Sparse Gaussion Process (SGP) to predict target score $y(m)$ given its latent vector. After training SGP, five iterations of batched Bayesian optimization (BO) are performed with expected improvement heuristics.

We report SGP prediction performance when trained on latent representations learned by different models. We train the SGP with 10-fold cross validation considering randomness and report the top-3 molecules found by the BO.

As shown in Table \ref{tab:BO}, molecules found by our model are much better than that by previous SMILES-based methods, and our method is even superior to the graph-based method JT-VAE. Figure \ref{fig:top3bo} shows top-3 molecules found by our model.

\section{Discussion}
Our method works extremely well in the molecule generation, in which SMILES sequences are highly structured and grammarly organized. Our experimental results confirm that grammar and syntax rules are necessary to generate more valid SMILES sequences. Besides, SMILES-based methods and graph-based methods may be combined together to boost the model performance further.

\bibliography{druggeneration.bib}
\bibliographystyle{aaai}

\end{document}